# An Architecture for Probabilistic Concept-Based Information Retrieval*

*Robert M. Fung, Stuart L. Crawford, Lee A. Appelbaum and Richard M. Tong*

Advanced Decision Systems
1500 Plymouth Street
Mountain View, California 94043-1230

## Abstract

While concept-based methods for information retrieval can provide improved performance over more conventional techniques, they require large amounts of effort to acquire the concepts and their qualitative and quantitative relationships.

This paper discusses an architecture for probabilistic concept-based information retrieval which addresses the knowledge acquisition problem. The architecture makes use of the probabilistic networks technology for representing and reasoning about concepts and includes a knowledge acquisition component which partially automates the construction of concept knowledge bases from data.

We describe two experiments that apply the architecture to the task of retrieving documents about terrorism from a set of documents from the Reuters news service. The experiments provide positive evidence that the architecture design is feasible and that there are advantages to concept-based methods.

## 1 Introduction

In this paper we describe some preliminary research on the use of probabilistic networks for information retrieval. In particular, we introduce an architecture

*This work was funded by ADS' Internal Research and Development Program.

for probabilistic, concept-based information retrieval (henceforth PCIR) that can be used first to automatically generate relationships between concepts and then reason about them given the evidence provided by individual documents. As in our previous research on concept-based methods [12, 16, 17], our goal has been to develop techniques that can be used to support a specific class of information retrieval problems. Specifically, we believe that the ideas we present here can form the basis for an effective system to assist users in sorting through large volumes of time sensitive material. We have in mind such applications as the day-to-day monitoring of newswires for specific topics of interest.

The architecture of a generic concept-based system is shown in Figure 1. A knowledge base contains a set of concepts together with their qualitative (*i.e.*, structural) and quantitative relationships with other concepts. Queries specify a user's information need in terms of these concepts. When a new document is presented with respect to a particular query, features are extracted from the document. The features currently used are the presence or absence of certain key words, and these features constitute evidence for the presence of concepts in the document. Using the features extracted from the document and the system knowledge base, inference is performed to assess the impact of the evidence on the belief in the query concept. The documents are sorted by belief and retrieved by a user-specified rule (*e.g.*, retrieve the "best" ten).

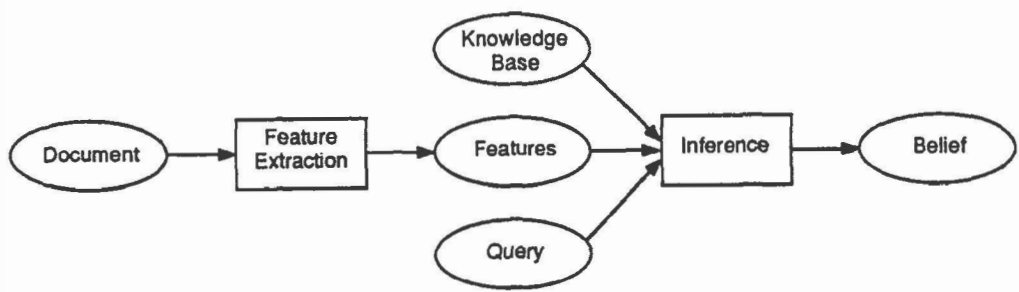

Figure 1: *Generic Concept-Based Architecture for IR*

Thus, concept-based methods view information retrieval primarily as a problem of evidential reasoning. However, while they can provide improved performance over more conventional techniques, they do require large amounts of effort to acquire the concepts and their relationships. Our current research attempts to address this weakness with the use of new probabilistic methods to represent, reason about, and learn the relationships between concepts. While probabilistic methods have been recognized as an important evidential reasoning technology with well-defined semantics (*e.g.*, frequency, strength-of-belief) and solid theoretical foundations, they have often been passed over because of their computational complexity. Their use in information retrieval has also been limited, although many authors have recognized the benefits of employing such techniques [2, 11, 14].

The probabilistic network technology [7, 13] is a recent development which is computationally tractable. A probabilistic network is a graph of nodes and arcs where the nodes represent uncertain variables and the arcs represent relationships between the variables. Computationally efficient algorithms have been developed which perform inference. The technology has been applied to a wide variety of problems including medical diagnosis, machine vision, petroleum exploration, military situation assessment, and multi-target tracking. Some initial work has applied this technology to information retrieval in hypertext [3, 5].

Because of the clear semantics behind probabilistic networks, it is possible to identify and quantify relationships between variables through experience (*i.e.*, data). CONSTRUCTOR [6] is a system for building probabilistic networks from data. It serves as the primary mechanism for learning about the relationship between concepts.

In the following section of the paper, we discuss both probabilistic networks and the CONSTRUCTOR system in more detail, and then, in Section 3, we describe the PCIR architecture. In Sections 4 and 5, we present the results of two exploratory experiments that show how we might use these techniques for concept-based retrieval. We conclude, in Section 6, with some comments and conclusions on the utility of the ideas we have presented.

## 2 Component Technologies

The two major component technologies of PCIR are probabilistic networks and CONSTRUCTOR.

### 2.1 Probabilistic Networks

Probabilistic networks is a technology for representing and reasoning with uncertain beliefs, and is based on the well-established theory of Bayesian probability. A successor to decision tree technology, probabilistic networks have been shown to be to be more understandable and computationally more tractable than the older technology. These advantages are achieved primarily through one innovation: the explicit representation of relevance relations between factors modeled in the network.

Table 1: $p(shoot|kill)$

|  | ¬shoot | shoot |
|---|---|---|
| ¬kill | 0.9 | 0.1 |
| kill | 0.1 | 0.9 |

There are two major types of probabilistic networks, Bayesian networks which contain directed arcs and Markov networks which contain undirected arcs. Both types are used in PCIR. There are two types of nodes: state and evidence nodes. A state node represents a mutually exclusive and collectively exhaustive set of propositions about which there is uncertainty. A state node is represented graphically by a circle. For example, whether a document is or is not about terrorism may be uncertain. To model this situation, the two propositions "this document is about terrorism" and "this document is not about terrorism" could be represented by a state node in a probabilistic network. An evidence node represents an observation and is represented graphically with a rectangle. For example, the observation that the word "bombing" is contained as a document may be represented as a evidence node in a probabilistic network.

Relationships between nodes in probabilistic networks are indicated with arcs. In a Bayesian network, a node's relationship with its *predecessors*[1] is what is modeled in a probabilistic network. Each node contains a probabilistic model of what is expected given every combination of predecessor values. For example, the predecessor of the *shooting* node in Figure 2 is the *killing* node. The probabilistic model for the *shooting* node is shown in Table 1. The model can be interpreted as saying that when the concept **killing** is present in a document, the concept **shooting** will probably be in the document and that when the concept **killing** is not present in a document then the concept **shooting** will probably not be found in the document.

---

[1] The set of nodes which have an arc which points to a given node are that node's predecessors.

In a Markov network, relationships between nodes are also indicated with arcs but represented in a different way. Probabilistic models are associated with the cliques (*i.e.*, maximally connected subset) of a network instead of individual nodes.

Relevance relations are specified by the connectivity of the network—what arcs are placed between what nodes, and in what direction. The concept of relevance in a Bayesian network is related loosely with graph separation and can be illustrated by examination of Figure 2. If it is known that the concept **killing** is present in a document, then the structure of the network implies that any other known information (*e.g.*, the concept **terrorism** is present in the document) will be irrelevant to beliefs about whether the **shooting** is present in the document. This is because the node *killing* separates the node *shooting* from every other node in the graph. Similarly, if it is known that the concepts **politician** and **terrorism** are both present in a document then any other piece of known information is irrelevant to whether the concept **subject** is present in the document. These relevance relations are useful not only from a qualitative point of view, but are also useful in reducing the amount of quantitative information needed and the amount of computational resources needed in inference.

Useful inferences can be made given a probabilistic network that represents a situation and evidence about the situation. For example, given the network representing the terrorism query and the evidence (*i.e.*, extracted features) from a document, one can infer an updated belief that the document is about terrorism. Several techniques are available for making inferences (*i.e.*, reaching conclusions) from a network and evidence. Shachter [15], Pearl [13], and Lauritzen and Spiegelhalter [10] all describe approaches to inference with probabilistic networks. Each approach has its advantages and disadvantages. For this work, we used the distributed algorithm [1, 9].

## 2.2 CONSTRUCTOR

The CONSTRUCTOR system [6] induces discrete, probabilistic models from data. These models contain a quantitative (*i.e.*, probabilistic) characteriza-tion of the data but, perhaps more importantly, also contain a qualitative structural description of the data. By qualitative structure we mean, loosely, the positive and negative *causal* relationships between factors as well as the positive and negative *correlative* relationships between factors in the processes under analysis. CONSTRUCTOR has as a primary focus the recovery of qualitative structures since structure not only determines which quantitative relationships are recovered, but also because such structure have been found to be cognitively stable [8] and thus are valuable in explaining the real world processes under analysis.

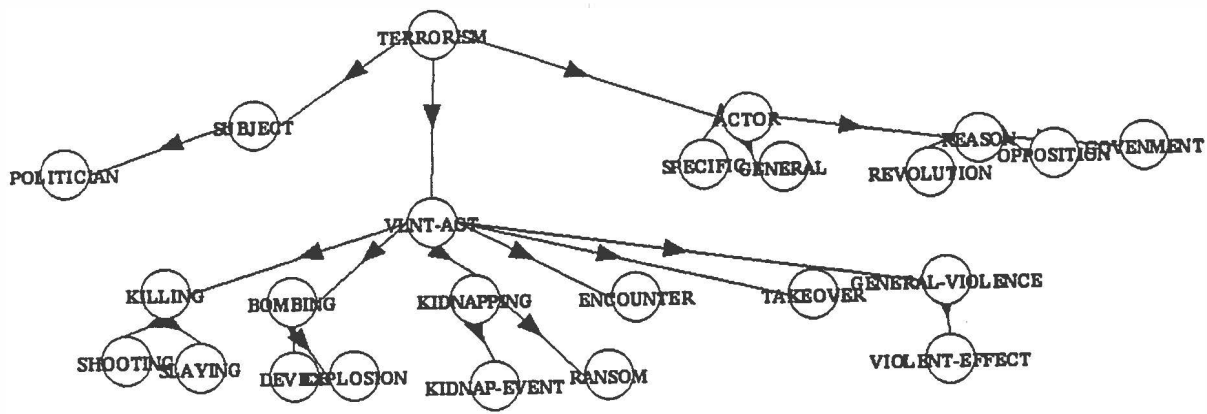

Figure 2: *Probabilistic Network for Terrorism*

The CONSTRUCTOR system is built upon techniques and research results from the fields of probabilistic networks, artificial intelligence (AI), and statistics. The probabilistic network technologies are central to the CONSTRUCTOR system since they not only provide the representation language for CONSTRUCTOR results but, more importantly, provide the conceptual impetus—the identification of conditional independence relations—that drives the CONSTRUCTOR system.

From the field of AI we have made use of heuristic search methods. These methods provide the primary problem solving paradigm of CONSTRUCTOR and allow for a computationally efficient implementation. From classical statistics, we make use of the $\chi^2$ test for probabilistic independence and from the newer field of computer-intensive statistical analysis [4] we make use of *cross-validation* to prevent "overfitting" of models to data.

The CONSTRUCTOR algorithm works by finding the complete set of (graphical) neighbors for each feature in the data set. The neighbor relations for each feature can then be used to identify the structure of a belief network. The complete set of neighbors for a feature is called the Markov boundary. The neighbors are identifiable as the smallest set of features such that all other features are conditionally independent of that feature given any fixed set of values for the feature's neighbors.

Network identification involves successively finding the *neighbors* of each attribute in the training set. Despite these observations, managing the exponential process of finding neighbors is the primary challenge for the network identification task. Finding the neighbors for every attribute in a training set is an iterative search process based on finding the Markov boundary for each attribute.

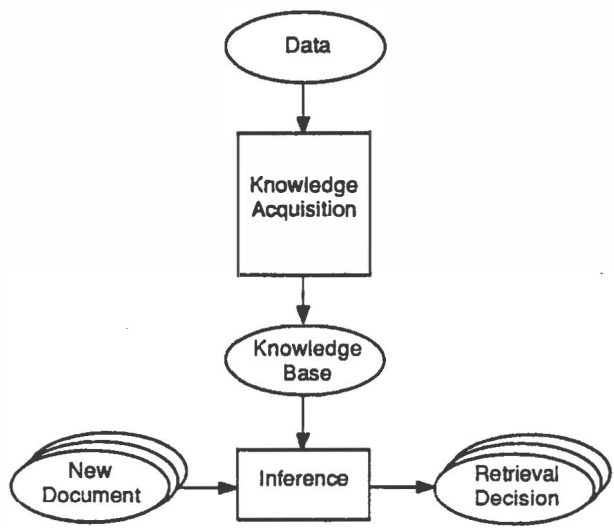

Figure 3: *PCIR Architecture*

## 3 Architecture

The PCIR architecture is shown in Figure 3. The major difference between it and the generic architecture in Figure 1 is the addition of the knowledge acquisition component. The rest of this section will discuss the PCIR knowledge base, the inference component and the knowledge acquisition component.

### 3.1 Knowledge Base

Central to the idea of a concept-based approach to information retrieval is a knowledge base which contains knowledge about relationships between concepts and features extractable from the document. In the PCIR architecture, the knowledge base takes the form of a set of probabilistic networks and can be obtained directly from a user or from the knowledge acquisition component of PCIR. The knowledge base consists of concept networks and concept-evidence relationships. A concept network relates concepts to other concepts. A concept-evidence relationship relates a concept to a subset of the features that will be extracted.

For example, the terrorism concept network shown in Figure 2 contains 24 concepts and requires the specification of 47 quantitative parameters. Also included in the knowledge base are 61 concept-evidence relationships which require the specification of an additional 64 parameters. The relationships encoded by both sets of parameters are intuitive and include:

- If the concept **terrorism** is in a document, it is almost twice as likely that the concept **violent act** will be in the document compared with the case that the document does not contain the concept **terrorism**.
- If the concept **bombing** is in a document, it is nine times as likely that the concept **explosion** will be in the document compared with the case that the document does not contain the concept **bombing**.
- If the concept **explosion** is in a document, it is four times as likely that the word "explosion" will occur in the document compared with the case that the document does not contain the concept **explosion**.

### 3.2 Inference

Given a document, a concept of interest and some decision criteria, the function of the inference component is to use the knowledge base created by the knowledge acquisition component to first judge the likelihood that the document contains the concept of interest and secondly to use that likelihood and the decision criteria to make a decision about retrieval of the document. Figure 4 shows the functional flow for the inference component.

The first step in the inference process is to extract a set of features from the document. Each feature must have values which are well-defined and must be mutually exclusive and exhaustive. The features currently used by PCIR are words that have been deemed to be relevant (by a PCIR user) to the set of concepts in the PCIR knowledge base. For example, the words "explosive" "blast" and "explosion" would likely be deemed relevant to the concept **explosion**. The features currently used in PCIR are binary-valued. The values represent whether or not a particular word is present or absent in a document.

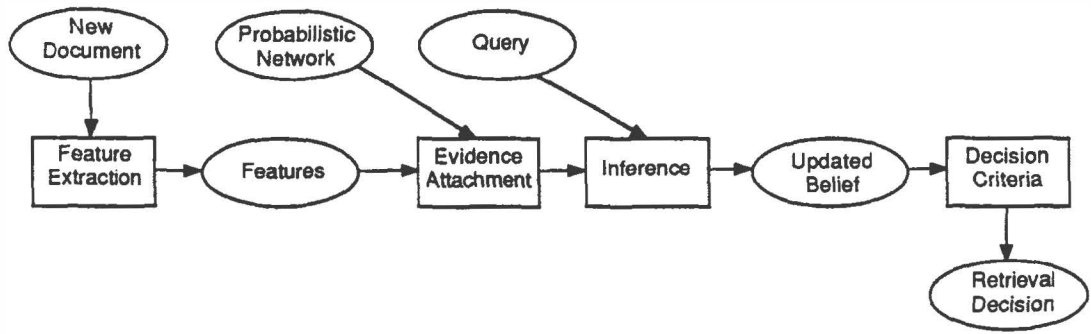

Figure 4: *PCIR Inference*

The result of feature extraction is a set of feature values. These feature values are instantiated as evidence nodes in the PCIR network and are attached to the appropriate state nodes (*i.e.*, concepts) in the network. The likelihoods which are required for the evidence nodes are derived from the concept-evidence relationships stored in the knowledge base.

The third step of the process is to perform probabilistic inference on the modified network. Given a concept of interest, the inference process computes the posterior distribution (*i.e.*, updated belief) of the concept given the evidence (*i.e.*, feature values) in the network. Since the concept of interest can be any of concepts in the network, a single network can serve to answer many queries.

The fourth step of the process is to apply the given decision criteria to the updated belief that the concept of interest is in the document. The decision criteria may be a simple threshold or may require comparison with the beliefs from other documents (*e.g.*, best n).

The probabilistic networks technology provides a probabilistic, model-based approach to deriving the strength of belief that a document contains a particular concept. By probabilistic, we mean that the domain knowledge of relationships between concepts and evidence is represented in probabilistic terms (*i.e.*, frequencies) and inference is performed with respect to the laws of probability. By model-based, we mean that the domain knowledge is represented as much as possible, in terms of behavioral models of cause and effect. For example the arc between the nodes *killing* and *shooting* in Figure 2 represents the belief that the presence of the concept **killing** in a document will with some probability, "cause" the presence of the concept **shooting**.

A model-based approach stands in contrast with an evidential-accrual approach, such as in RUBRIC [12, 16, 17]. The flow of reasoning in evidence-accrual approaches is directly from from effect to cause (*i.e.*, evidence to conclusions). Evidence is accrued to the first level of conclusions which in turn act as evidence for the next higher layer of conclusions. In contrast, the flow of reasoning in model-based reasoning approaches can be viewed as a two pass process. In the first pass, reasoning flows from cause to effect in order to set up expectations for the evidence. And in the second pass, these expectations are compared with the actual evidence, and the comparisons are transmitted back from the effects to the causes.

In applying belief networks to information retrieval, one major decision was required—what states should the nodes represent. We choose to follow RUBRIC by assigning two states to each node in a network, where the states represent that a concept is present or absent in a document. Given this choice of states the probability distributions of a network represent beliefs about how the presence of sets of concepts in a document "causes" or "correlates with" the presence of other concepts in the document. For example, the model shown in Figure 2 shows that the presence of the concept **terrorism** in a document "causes," to some (probabilistic) degree, the inclusion of the concept **terrorist actor** to be in the document.

### 3.3 Knowledge Acquisition

While concept-based approaches such as RUBRIC are able to provide good results, the effort needed to acquire the knowledge bases needed by such approaches from experts requires substantial resources. PCIR provides an approach to reducing the effort needed for knowledge acquisition.

Given a set of documents, a set of features, and a set of concepts, the function of the knowledge acquisition component is to develop a knowledge base which establishes relationships between concepts and features. Figure 5 shows the functional flow for the knowledge acquisition component.

The user of PCIR must provide the inputs to the knowledge acquisition component. The inputs are a set of documents, a set of features, and a set of concepts. The document set is a population of documents which should be representative of the documents which will be faced in retrieval. The Reuters document collection used to generate the terrorism network contains 730 documents.

A set of concepts must be identified. Usually the concepts are identified through association (by the user) to the concept on which it is anticipated most retrievals will be performed. For example, the concepts included to generate the terrorism network were associated with the concept **terrorism**.

Given these inputs, there are two steps required to create a CONSTRUCTOR data set: feature extraction and concept specification.

Feature extraction is exactly the same process as in the inference component and is performed for each document in the document set.

Concept specification is the most user-intensive process in the architecture. The user must specify for each document in the data set which of the concepts in the concept specification are contained in the document.

Appending the concept specification and the feature values for each document creates a data set which can be processed by CONSTRUCTOR. The data set consists of an array of values. Each row represents the concepts and features present in a particular document. Each column represents a particular feature or concept.

The result of processing the data set through CONSTRUCTOR is a probabilistic network that can act as the knowledge base for the inference component.

If the user desires, a threshold decision criteria can be obtained for a particular concept of interest by passing each of the documents through the inference component of PCIR. A threshold can then be chosen by the user which provides for an appropriate tradeoff between precision and recall.

## 4 Experiments

Two simple experiments were performed with the Reuters database. The first experiment entailed building a probabilistic network where both the structure of the network and the probability distributions were given by a "user" (the principal author). In the second experiment, a probabilistic network was built using the Reuters database as input to the knowledge acquisition component of PCIR.

### 4.1 "Hand-constructed" Network

A simple network was built around the **terrorism** concept, using as a model a RUBRIC concept tree built for **terrorism**. The network contains 23 concepts and was developed in a hierarchical fashion similar to the RUBRIC concept tree. The *terrorism* node was broken down into an *actor* performing a *violent act* on some *subject*. Similarly, *violent act* was broken down into different types of violent acts etc.. This network required 47 independent probability assessments. Except for the prior distribution on the *terrorism* node which was set to the frequency of terrorist documents in the document set, the probabilities were assessed qualitatively by the "user."

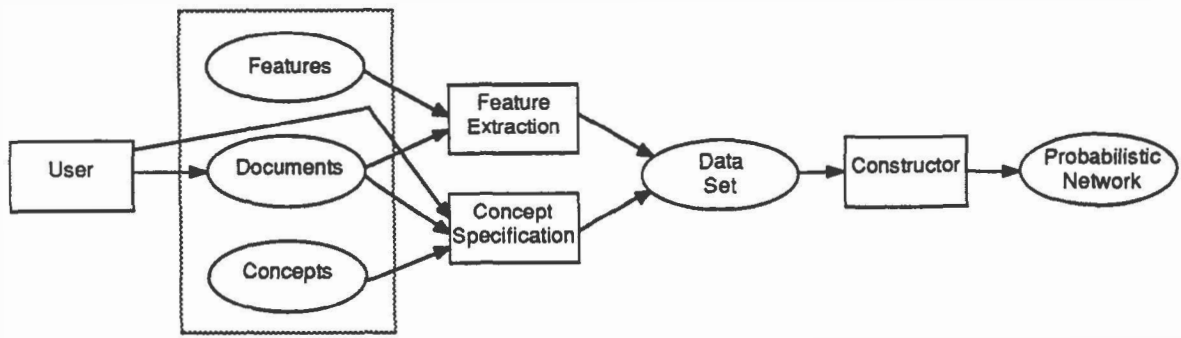

Figure 5: *PCIR Knowledge Acquisition*

A set of 61 features (*i.e.*, words) were extracted from each document. Each of these features requires a concept-evidence relationship to be present in the knowledge base. (These relationships are not shown.) Each of the 61 words were assigned by the user to a single concept and probabilities were specified for the events that a word appears in a document given their assigned concept is present in a document.

Because of the difficulty of the knowledge acquisition task, several assumptions were made to reduce the number of parameters needed to be specified for this network. The assumptions included the hierarchical structure of the network as well as a constant likelihood that a word does not appear given its assigned concept does not appear in a document. The latter assumption effects the posterior probabilities so that they are not "normalized." However the assumption does not effect the separation of the populations.

Using the concept-evidence relationships, evidence was attached to the probabilistic network in Figure 2. The mean and standard deviation of the posterior probability for both documents about terrorism and documents not about terrorism is shown in Table 2. It can be seen that the posterior probability of documents about terrorism is significantly higher than for the documents not about terrorism.

The precision and recall results for a range of possible thresholds are shown in Figure 6. In the middle range both precision and recall are approximately 50%. While RUBRIC results are significantly better, much less effort was expended on this experiment and the results are competitive with conventional techniques.

The goal of the experiment was to assess the feasibility of using probabilistic networks as the evidential reasoning mechanism in a concept-based information retrieval scheme. This experiment seems to suggest that this is feasible. Some effort was made to see if some parameter modification might easily improve performance. To this end the feature sets of relevant, unretrieved documents and irrelevant, retrieved documents were examined. While several modifications where made, no significant performance improvements were found. This points out the difficulty of knowledge acquisition from experts, not only in the initial acquisition stage but also in the knowledge base tuning stage.

Table 2: *"Hand-constructed" results*

|  | avg | std dev |
|---|---|---|
| **terrorism** | .035 | .03 |
| ¬**terrorism** | .015 | .008 |

Table 3: Concepts

| unnamed terrorist | named terrorist | assassination |
|---|---|---|
| politician | government | opposition |
| reason | takeover | encounter |
| kidnap event | ransom | explosion |
| bombing | device | shooting |
| killing | violent act | violent effect |

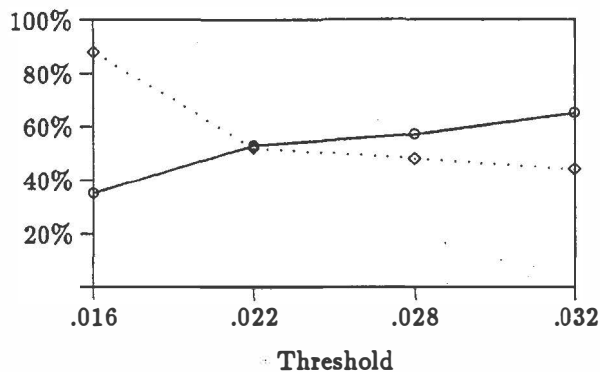

Figure 6: *Precision (solid line) & Recall (dotted line) vs. Threshold.*

### 4.2 Using CONSTRUCTOR

For this experiment, each of the 730 Reuters documents was tagged according to whether or not the document was "about" terrorism. Decisions about the relevancy of each document to the **terrorism** concept were made by a independent pair of readers. A total of 50 documents were judged to contain the concept **terrorism** and the other 680 were judged not to contain the concept **terrorism**. A set of 82 words was selected as the feature set. The presence or absence of each of the 82 words was determined for each document in the document set. In addition, 18 concepts were chosen as being possibly relevant to the concept of terrorism. The concepts are shown in Table 3. The two readers were also asked to indicate which of the 18 different concepts were relevant to each of the 730 Reuters documents.

CONSTRUCTOR was first run with a data set made up of the 18 concepts plus the **terrorism** concept. The resulting Markov network is shown in Figure 7. Nodes for which there is not a path to the *class* node (*i.e.*, terrorism) are not shown. Many of the arcs have intuitive interpretations which are supported by the underlying probability distributions found in the data.

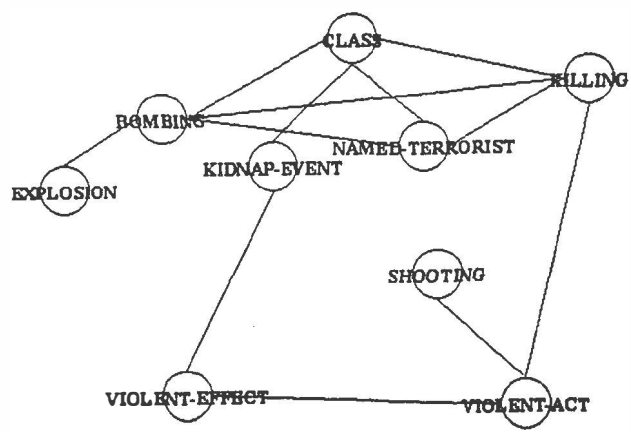

Figure 7: CONSTRUCTOR *Network for Terrorism*

- "A bombing causes an explosion."
- "A shooting is a violent act."
- "A killing is a violent act."
- "A terrorist event is present if two or more of the concepts bombing, named terrorist, killing or kidnapping is present except for the combination named terrorist and killing."

The concept-evidence relationships were derived for each of 8 concepts in the network, by running CONSTRUCTOR on a data set which included one of the concepts and the words associated with the concepts. Given these results, the knowledge base was complete.

To test the network's performance, each of the 730 documents was processed by the inference component of PCIR using the CONSTRUCTOR-derived knowledge base. The mean and standard deviation of the posterior probability for both documents about terrorism and documents not about terrorism is shown in Table 4. It can be seen that the posterior probability of documents about terrorism is significantly higher than for the documents not about terrorism and that separation of the populations is well-defined.

Table 4: CONSTRUCTOR results

|  | avg | std dev |
|---|---|---|
| terrorism | .45 | .21 |
| ¬terrorism | .036 | .09 |

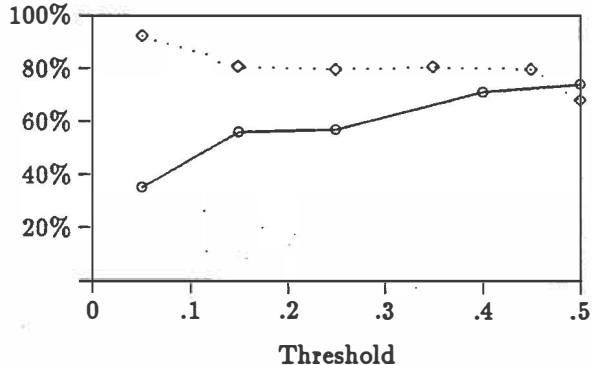

Figure 8: *Precision (solid line) & Recall (dotted line) vs. Threshold.*

Table 5: $p(explosion|bombing)$

|  | ¬explosion | explosion |
|---|---|---|
| ¬bombing | 0.98 | 0.02 |
| bombing | 0.59 | 0.41 |

Table 6: $p(terrorist|terrorism)$

|  | ¬terrorist | terrorist |
|---|---|---|
| ¬terrorism | 0.98 | 0.02 |
| terrorism | 0.64 | 0.36 |

The precision and recall results for a range of possible thresholds are shown in Figure 8. In the middle range both precision and recall are in the 70% to 80% range.

As this was the first application of CONSTRUCTOR to real data, we were pleased with the robustness and intuitiveness of the relationships and the performance of the resulting network.

Many of the relationships that were found are quite robust and had similar structures. Consider Table 5, and Table 6 as examples. In both these tables, the relations between the nodes can be interpreted as noisy if-then statements:

- "If the concept **bombing** is *not* in a document, then the concept **explosion** will *not* be in the document"
- "If the concept **terrorism** is *not* in a document, then the concept **terrorist** will *not* be in the document"

On the other hand, the contrapositive versions of these statements which are perhaps more intuitive, are not true. It is not true that:

- "If the concept **bombing** is in a document, then the concept **explosion** will be in the document" or
- "If the concept **terrorism** is in a document, then the concept **terrorist** will be in the document"

While many of the structural relationships and their corresponding quantitative relationships in the network are intuitive, there are some complicated relationships present in the network which are quite subtle. For example, consider the relationship between the concept **killing** and the concept **terrorism**. Whereas the other neighbors of **terrorism** (*i.e.*, **bombing, kidnap,** and **named-terrorist**) have strong, uncomplicated relationships with **terrorism**, the concept **killing** seems to have a relatively small effect by itself but seems to act as a magnifier of the positive influence of the other neighbors. This can be seen in Table 7 and was borne out when the frequencies of these events were examined in the raw data. Such subtle relationships may be the cause of the CONSTRUCTOR network's improved performance over the "hand-constructed" network and it is easy to imagine that such relationships would take much effort to find manually.

Table 7: $p(terrorism|bombing, kidnap, killing, terrorist)$

|  |  |  |  | ¬terrorism | terrorism |
|---|---|---|---|---|---|
| ¬bombing | ¬kidnap | ¬killing | ¬terrorist | 0.994 | 0.006 |
| ¬bombing | ¬kidnap | ¬killing | terrorist | 0.85 | 0.15 |
| ¬bombing | ¬kidnap | killing | ¬terrorist | 0.96 | 0.04 |
| ¬bombing | kidnap | ¬killing | ¬terrorist | 0.61 | 0.39 |
| bombing | ¬kidnap | ¬killing | ¬terrorist | 0.68 | 0.32 |
| bombing | kidnap | ¬killing | ¬terrorist | 0.02 | 0.98 |
| bombing | ¬kidnap | killing | ¬terrorist | 0.44 | 0.56 |
| ¬bombing | kidnap | killing | ¬terrorist | 0.51 | 0.49 |
| bombing | ¬kidnap | ¬killing | terrorist | 0.24 | 0.76 |
| ¬bombing | kidnap | ¬killing | terrorist | 0.05 | 0.95 |
| ¬bombing | ¬kidnap | killing | terrorist | 0.8 | 0.2 |
| bombing | ¬kidnap | killing | terrorist | 0.53 | 0.47 |
| ¬bombing | kidnap | killing | terrorist | 0.07 | 0.93 |
| bombing | kidnap | ¬killing | terrorist | 0.003 | 0.997 |
| bombing | kidnap | killing | ¬terrorist | 0.03 | 0.97 |
| bombing | kidnap | killing | terrorist | 0.02 | 0.98 |

## 5 Conclusions

We believe that the experimental results presented above provide positive evidence that the PCIR architecture design is feasible. The choice of probabilistic networks for the knowledge base representation provides for an intuitive and well-defined semantics for acquiring knowledge either from an expert or automatically. The first experiment shows that reasonable performance can be obtained through use of probabilistic networks as the evidential reasoning mechanism for concept-based information retrieval. The second experiment reinforces this conclusion while also showing that partially automating the knowledge acquisition task is possible.

The central hypothesis of concept-based methods for information retrieval is that the representation of, and reasoning about, unobservable concepts is effective both from an organizational and from a computational point of view. We feel that a secondary contribution of this work is positive evidence for this hypothesis. All the evidence stems from the assumption that the CONSTRUCTOR-induced network is close to being correct and the fact that the network is sparse (i.e., has few arcs). Out of the 153 possible arcs between the 18 concept nodes of the graph, only 12 of the arcs are instantiated. In addition, there are the 82 connections to the 82 evidence (e.g., feature) nodes. In contrast, consider the situation if all the concept nodes except the terrorism node were removed from the graph by probabilistic manipulation. The resulting graph would be extremely dense. This would correspond to the situation of deriving probabilistic relations between terrorism and the features directly.

Three advantages for concept-based methods can be seen from this analysis. First, concepts organize information into a small number of manageable concept-to-concept and concept-to-feature relations. This makes both manual and automatic knowledge acquisition easier. Secondly, concepts reduce the computational complexity of inference. Probabilistic inference is inherently easier in sparse networks than in dense networks. Thirdly, concepts make the automatic knowledge acquisition problem tractable by dramatically reducing the sampling problem. The probability tables of dense networks are exponentially larger than the probability tables for sparse networks. Dense networks will therefore spread the examples in the training set over a much larger space. Undersampling can be a serious problem in such situations. On the other hand, sparse networks do not suffer from such problems.

As a secondary point, we feel that not only are concepts useful computationally, but the robustness of the relationships between concepts seen in the CONSTRUCTOR-induced network provides strong evidence for the psychological intuition that these concepts are cognitively significant in people's thought processes.

The most visible drawback of this research is the amount of work needed by a user to identify what concepts are present for each document in a large document set. However, we think a scenario in which a user incrementally performed this is certainly feasible. Also, if the concepts of interest are not in a special domain, this work can be done by relatively untrained people. A research goal is to identify concepts automatically by clustering.

We think that the results are promising and intend to pursue further research in this direction. Further experimentation with the Reuters document set and the terrorism query is planned. Another area of research is experimentation with different document sets, different features, and different concepts. The CONSTRUCTOR algorithm itself is new and evolving. Improvements to the algorithm could be the source of important improvements to PCIR.

# References

[1] K. C. Chang and R. M. Fung. Node aggregation for distributed inference in bayesian networks. In *Proceedings of the 11th IJCAI*, Detroit, Michigan, August 1989.

[2] W. B. Croft and D. J. Harper. Using probabilistic models of document retrieval without relevance information. *Journal of Documentation*, 35:285–295, 1979.

[3] W. B. Croft and H. Turtle. A retrieval model incorporating hypertext links. In *Hypertext'89 Proceedings*, November 1989.

[4] B. Efron. Computers and the theory of statistics: thinking the unthinkable. *SIAM Rev-21*, 460–480, 1979.

[5] M. E. Frisse and S. B. Cousins. Information retrieval from hypertext: update on the dynamic medical handbook project. In *Hypertext'89 Proceedings*, November 1989.

[6] R. M. Fung and S. L. Crawford. Constructor: empirical acquistion of probabilistic models. In *Seventh International Conference on Machine Learning*, June 1990. submitted.

[7] R. A. Howard and J. E. Matheson. Influence diagrams. In R.A. Howard and J.E. Matheson, editors, *The Principles and Applications of Decision Analysis, vol. II*, Menlo Park: Strategic Decisions Group, 1981.

[8] D. Kahneman, P. Slovic, and A. Tversky. *Judgement under uncertainty: Heuristics and biases*. Cambridge University Press, Cambridge, 1982.

[9] J. H. Kim and J. Pearl. A computational model for combined causal and diagnostic reasoning in inference systems. In *Proceedings of the 8th Internationl Joint Conference on Artificial Intelligence*, Los Angeles, California, 1985.

[10] S. L. Lauritzen and D. J. Spiegelhalter. Local computations with probabilities on graphical structures and their application in expert systems. *Journal Royal Statistical Society B*, 50, 1988.

[11] M. E. Maron and J. L. Kuhns. On relevance, probabilistic indexing and information retrieval. *Journal of the ACM*, 7:216–244, 1960.

[12] B. P. McCune, R. M. Tong, J. S. Dean, and D. G. Shapiro. RUBRIC: a system for rule-based information retrieval. *IEEE Transactions on Software Engineering*, SE-11(9):939–945, 1985.

[13] J. Pearl. *Probabilistic Reasoning in Intelligent Systems: Networks of Plausible Inference*. Morgan Kaufmann Publishers, 1988.

[14] S. E. Robertson, C. J. van Rijsbergen, and M. F. Parker. Probabilistic models of indexing and searching. In R. N. Oddy, S. E. Robertson, C. J. van Rijsbergen, and P. N. Williams, editors, *Information Retrieval Research*, London: Butterworths, 1981.

[15] R. D. Shachter. Intelligent probabilistic inference. In L.N. Kanal and J.F. Lemmer, editors, *Uncertainty in Artificial Intelligence*, Amsterdam: North-Holland, 1986.

[16] R. M. Tong, L. A. Appelbaum, and V. N. Askman. A knowledge representation for conceptual information retrieval. *Int. J. Intelligent Systems*, 4(3):259–284, 1989.

[17] R. M. Tong and D. G. Shapiro. Experimental investigations of uncertainty in a rule-based system for information retrieval. *Int. J. Man-Machine Studies*, 22:265–282, 1985.